\pdfoutput=1
\documentclass[runningheads]{llncs}
\usepackage{float}
\usepackage{graphicx}
\usepackage{amsmath}
\usepackage{amsfonts}
\usepackage{multirow}
\usepackage{booktabs}
\usepackage{arydshln}
\usepackage{placeins}
\usepackage{marvosym}
\begin{document}
\title{RefineStyle: Dynamic Convolution Refinement for StyleGAN}
\titlerunning{RefineStyle: Dynamic Convolution Refinement for StyleGAN}
%
\author{Siwei~Xia \and
Xueqi~Hu \and
Li~Sun\textsuperscript{(\Letter)} \and
Qingli~Li}
\authorrunning{S.~Xia et al.}
\institute{Shanghai Key Laboratory of Multidimensional Information Processing, East China Normal University, Shanghai, China\\
\email{sunli@ee.ecnu.edu.cn}
}
\maketitle              
\renewcommand{\thefootnote}{}
\footnotetext{S. Xia and X. Hu—These authors contributed equally to this work.}
\begin{abstract}
In StyleGAN, convolution kernels are shaped by both static parameters shared across images and dynamic modulation factors $w^+\in\mathcal{W}^+$ specific to each image. Therefore, $\mathcal{W}^+$ space is often used for image inversion and editing. However, pre-trained model struggles with synthesizing out-of-domain images due to the limited capabilities of $\mathcal{W}^+$ and its resultant kernels, necessitating full fine-tuning or adaptation through a complex hypernetwork.
This paper proposes an efficient refining strategy for dynamic kernels. The key idea is to modify kernels by low-rank residuals, learned from input image or domain guidance. These residuals are generated by matrix multiplication between two sets of tokens with the same number, which controls the complexity. We validate the refining scheme in image inversion and domain adaptation. 
In the former task, we design grouped transformer blocks to learn these token sets by one- or two-stage training. 
In the latter task, token sets are directly optimized to support synthesis in the target domain while preserving original content. Extensive experiments show that our method achieves low distortions for image inversion and high quality for out-of-domain editing.

\end{abstract}

\section{Introduction}
Given a noise vector from a prior distribution, Generative Adversarial Network (GAN) has shown strong ability to output high fidelity images. StyleGAN\cite{karras2019style,karras2020analyzing,karras2021alias}, a leading GAN variant, is expected to play a pivotal role in downstream tasks, like image editing, image-to-image translations, \emph{etc.} 
To take full advantage of pre-trained StyleGAN, the key issue is to obtain out-of-domain image from its output, which is challenging for image inversion and domain adaptation. Specifically, we aim to find an optimal latent space within the pre-trained generator to faithfully reproduce a given image or a particular style. 

Based on dynamic modulation design in StyleGAN, researchers have investigated different latent spaces. It is found that the initial noise space $\mathcal{Z}$ is inferior to the intermediate space $\mathcal{W}$~\cite{abdal2019image2stylegan}, in which style code $w\in \mathcal{W}$, a single vector of 512-d, modulates the convolution kernels. Due to style mixing during training, $\mathcal{W}$ can be further extended to $\mathcal{W}^+$, giving independent control on each convolution layer. The code $w^+\in\mathcal{W}^+$ consisting of $18$ different 512-d vectors, is more suitable for high quality inversion and editing. While most of existing methods~\cite{abdal2019image2stylegan,richardson2021encoding} use $\mathcal{W}^+$, challenges persist in accurately capturing unique appearances, \emph{e.g.}, hat, heavy makeup, or different domain style. To complement the $\mathcal{W}^+$ space, researchers propose to fully finetune~\cite{roich2021pivotal,gal2021stylegan} the generator or utilize hypernetwork~\cite{alaluf2022hyperstyle,dinh2022hyperinverter}. Refining the generator's parameters is able to accommodate the style variance, thus achieves the high fidelity out-of-domain synthesis. Whereas, the numerous parameters increase the training complexity and costs.

To enhance out-of-domain synthesis without high costs, we proposes an efficient low-rank refining scheme for dynamic convolution kernels in StyleGAN. In the computation graph, $w^+$ code, specified by the side branch MLP, determines the modulation factors on static parameters in the backbone, thus dynamically forming convolution kernels. Compared to $w^+$, these dynamic kernels directly control synthesis, so adapting them for inversion and adaptation is reasonable. To refine them, residuals should be simple and small, hence they don't affect original StyleGAN severely. Accordingly, they are designed to be low-rank with learnable scaling factors to modulate their influence. Specifically, we employ two token sets for each StyleGAN layer, matching the input and output channels of dynamic convolution layer respectively. Each regarded as a matrix, the two sets are multiplied to produce low-rank residuals in the same shape with kernels.

For image inversion, we design grouped transformer blocks to learn the desired token sets from each image. Within these blocks, tokens interact exclusively with others from the same group in self-attention, while in cross-attention, all tokens engage with features from the original image. Based on these blocks, we can build either a two-stage model, which first computes $w^+$ then refines dynamic kernels, or a one-stage model that handles both simultaneously.

For domain adaptation, token sets are designed to reflect styles from target texts or reference images. Thus we simply optimize them towards specific domain defined by text descriptions as in StyleGAN-NADA~\cite{gal2021stylegan} or by the reference image as in GOGA~\cite{Zhang2022GeneralizedOD}, with parameters of the original StyleGAN intact. Due to the effective low-rank assumption, synthesis achieves high quality in the target domain.

This paper's contributions are summarized as follows: 
\begin{itemize}
  \item We propose a refining scheme for the dynamic kernels in pre-trained StyleGAN. Our method learns two token sets for each layer of the generator, and multiplies them to obtain low-rank residuals that augment the kernels.
  \item We apply the proposed scheme to out-of-domain image inversion. We design grouped transformer blocks to learn desired tokens from the input image with either one-stage or two-stage training, allowing for faithful reconstruction of real data.
  \item To show its versatility, we utilize our proposed model in domain adaptation. The token sets are directly driven to give synthesis in the target domain, while keeping prior knowledge in the original domain.
\end{itemize}

\section{Related Works}
\textbf{Image inversion.} 
GAN inversion~\cite{zhu2016generative} aims at projecting real images into latent representations. Typically, inversion methods are categorized into two types: encoder-free and encoder-based.
Encoder-free methods, like I2S~\cite{abdal2019image2stylegan,abdal2020image2stylegan++}, directly optimizing $w^+\in\mathcal{W}^+$ to faithfully reconstruct images, however, lack editability. Encoder-based methods are more efficient and edit-friendly, but introduce larger distortions in out-of-domain images. Feature pyramid~\cite{richardson2021encoding,tov2021designing} was introduced to obtain multi-resolution features for enhancement. Transformer decoder with cross-attention module~\cite{hu2022style} interacting between learnable queries and image features proves to further improve synthesis. These methods operate within $\mathcal{W}^+$ space while some works introduce a second stage to tune the pre-trained generator $G$. HFGI~\cite{wang2022high} employs another encoder for further feature modulation post dynamic convolution. HyperStyle~\cite{alaluf2022hyperstyle} and HyperInverter~\cite{dinh2022hyperinverter} adapt dynamic convolution to various images through a hyper-network. CLCAE~\cite{liu2023delving} aligns $w\in\mathcal{W}$ and then designs an encoder for more codes. StyleRes~\cite{pehlivan2023styleres} learns residual features in higher latent codes.
These works, characterized as two-stage methods built on existing $\mathcal{W}^+$ inverters, markedly improve out-of-domain inversion. Our method can be implemented in one- or two-stage, and either achieves more accurate results.
\begin{figure}[!t]
\centering
\includegraphics[width=2.5in]{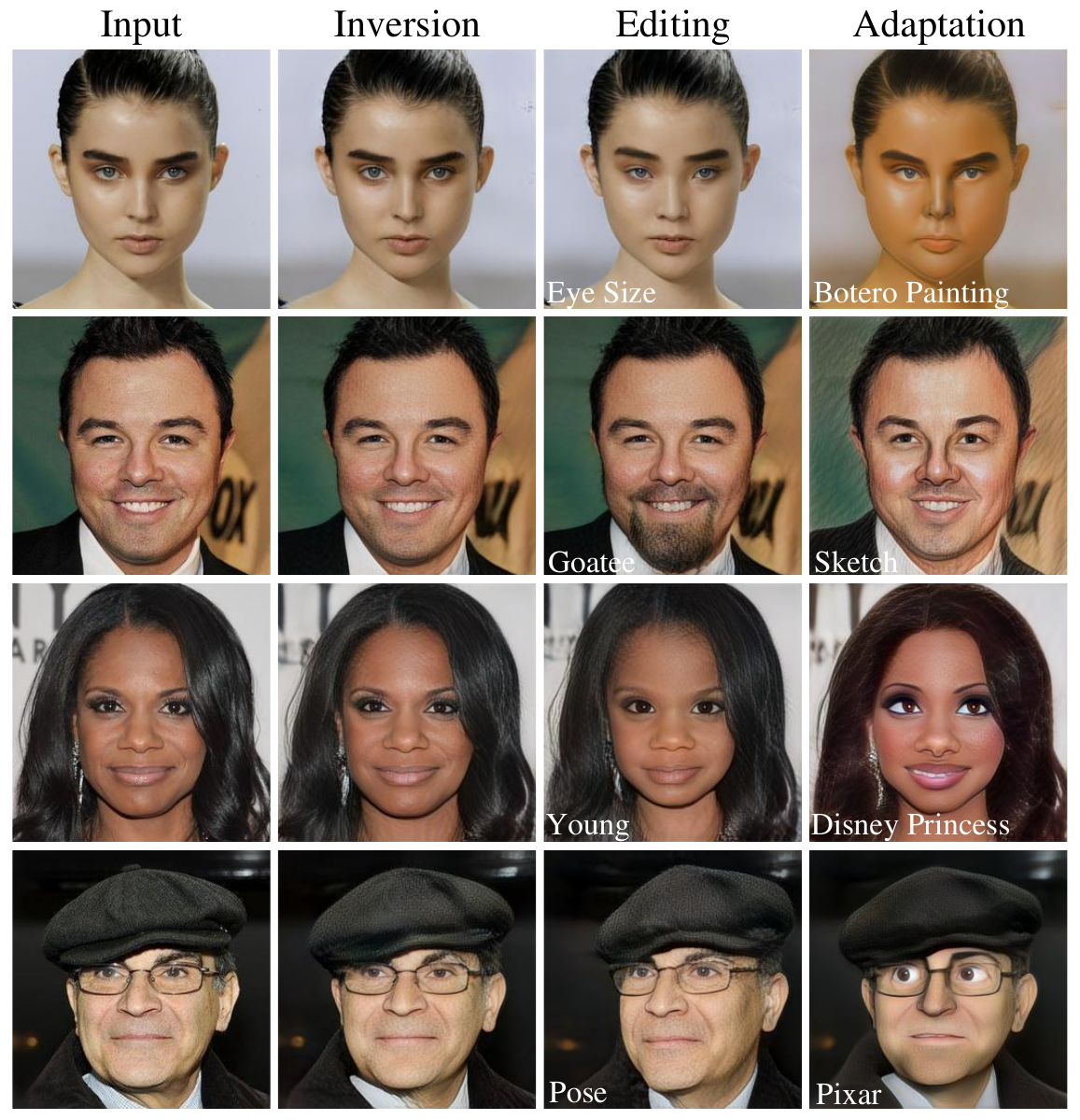}
\caption{Simulation results for RefineStyle.}
\label{fig_1}
\end{figure}

\textbf{Latent code editing.} 
Pre-trained StyleGAN has been widely adopted for in-domain image editing. Most methods first obtain the inverted latent code of original image, and then modify it to fulfill the requirement. InterFaceGAN~\cite{shen2020interpreting} employs linear classifier in the latent space to obtain editing directions. StyleTrans~\cite{hu2022style} adopts latent classifier for code optimization. StyleSpace~\cite{wu2021stylespace} detects the sensitive channel in $\mathcal{S}$ space for each facial attribute. These methods require images with labels. Besides, there are unsupervised methods, such as GANSpace~\cite{harkonen2020ganspace}, using PCA to determine editing directions. We utilize these existing methods for label-based editing.

\textbf{Domain adaptation.} 
Domain adaptation aims to fine-tune a pre-trained generator to a new domain under limited training data~\cite{wang2018transferring}. In~\cite{ojha2021few}, correspondences between source and target domain are set up to fine-tune the generator. StyleGAN-NADA~\cite{gal2021stylegan} proposes a direction loss to guide the generator based on textual descriptions. GOGA~\cite{Zhang2022GeneralizedOD} replicates the style of a single reference image across synthesized images. Instead of fully fine-tuning, GenDA~\cite{yang2021one} trains a simple adapter for both generator and discriminator. DualStyleGAN~\cite{yang2022Pastiche} adapts dynamic kernels at various resolutions with progressive multi-stage training. Our proposed RefineStyle shares the similar idea, but we emphasize that the low-rank residuals for modifying kernels are important to keep original content, and efficient to achieve target style without complex training strategies.

\begin{figure}[t]
\centering
\includegraphics[width=0.85\columnwidth]{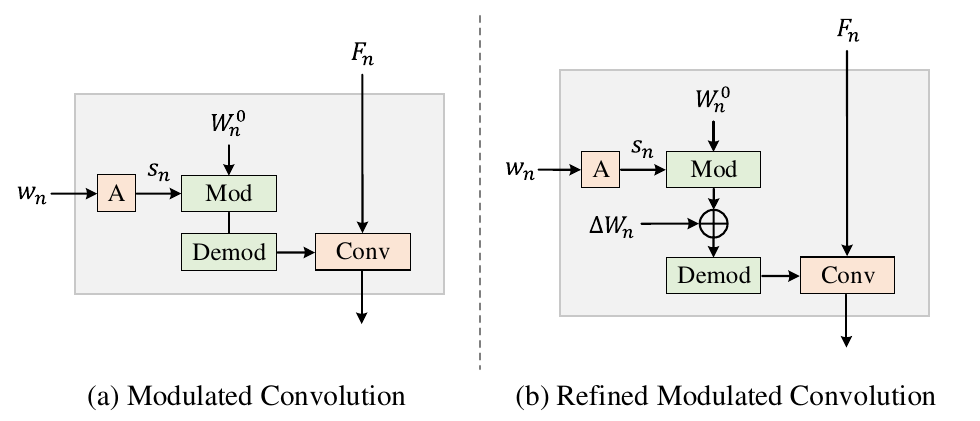}
\caption{(a) Modulated convolution in the n-th layer of StyleGAN2. (b) The proposed refined modulated convolution.}
\label{fig:structure}
\end{figure}

\section{Method}
\subsection{Problem Formulation}
We aim to take full use of pre-trained StyleGAN2 generator $G$ to synthesize images beyond its original training domain. Considering the computation flow, StyleGAN2 has several candidate latent spaces to describe images. The initial noise vector $z\in\mathbb{R}^{512}$ is first mapped into style code $w$ through an MLP, then projected to $s_n$ for modulating the kernel $W_n$ in $n$-th layer of $G$, where $n=1,2,\cdots, 17$ in StyleGAN2. The structure of modulated convolution is shown in Fig. \ref{fig:structure}(a), and the computation process is given in Eq. (\ref{eq:1}).
\begin{equation}
\label{eq:1}
    w=\text{MLP}(z)\quad s_n=\text{FC}_n(w)\quad
    W_n= s_n \odot W_n^0
\end{equation}
Here $W_n^0$ is the static kernel in pre-trained backbone $G$, $n$ denotes the layer index, and $\odot$ indicates element-wise product. In practice, $w^+\in \mathbb{R}^{18\times 512}$, comprised of 18 different $w$, is used to independently modulate each dynamic kernel $W_n$. 

Although $w^+$ is able to represent general contents of in-domain real images, it falls short in describing details, particularly for out-of-domain data. Recent works~\cite{roich2021pivotal,alaluf2022hyperstyle,wang2022high} either fine-tune the weights of pre-trained generator, or train a parametric adapter to refine the dynamic kernels. Our work belongs to the latter type, but we focus on finding simple and relevant modifications for $W_n$. 

\subsection{Refining Dynamic Kernels}
\subsubsection{Analysis on Kernel Space}
\begin{figure}[h]
\centering
\includegraphics[width=.5\columnwidth]{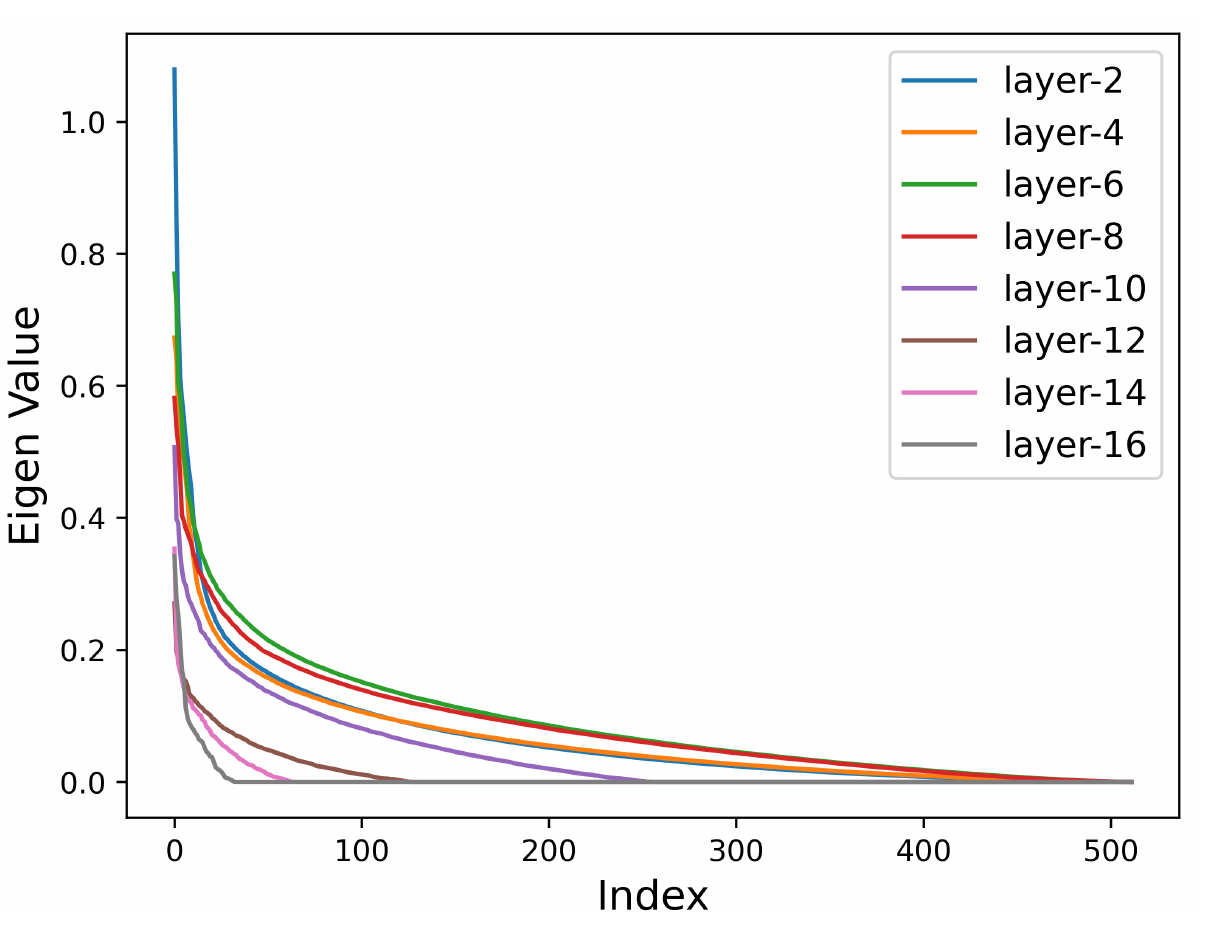}
\caption{Sorted eigen-values of dynamic kernels. For each layer, we perform SVD on kernel weight $W_n$, and list all eigen-values in the descending order. Notice that the first few values constitute most of the total, indicating that kernels are low-rank. }
\label{fig:svd}
\end{figure}

Expecting to understand the distribution of convolution kernel $W_n\in\mathcal{K}$ , we first randomly sample a number of $z$ from the standard Gaussian as the input of the MLP, and apply singular value decomposition (SVD) on the modulated kernel $W_n$ of each layer. As shown in Fig. \ref{fig:svd}, a few singular values significantly outweigh the rest in the same layer, indicating that the convolution weights lie in a low-dimensional manifold. This observation implies that refining $W_n$ with a low-rank residual is sufficient.

\subsubsection{Low-Rank Dynamic Residuals}
Since the high-dimensional kernel space $\mathcal{K}$ lies in a low dimensional manifold, we refine the kernels by learning low-rank residuals, as shown in Fig. \ref{fig:structure}(b). 
Formally, in the $n$-th layer, the static convolution kernel $W_n^0\in\mathbb{R}^{C_{out}\times C_{in} \times 3 \times 3}$ undergoes two dynamic modifications. The first is modulation by $s_n$ yielding $W_n$ as Eq.(\ref{eq:1}), and the second is addition of residual $\Delta W_n\in\mathbb{R}^{C_{out}\times C_{in}}$ to form the final kernel $W_n^d$ as Eq.(\ref{eq:2}). Note that since $\Delta W_n$ lacks spatial dimensions, we simply expand it to the same size as $W_n$.
\begin{equation}\label{eq:2}
    W_n^d = \Delta W_n \oplus  W_n = \Delta W_n \oplus (s_n\odot W_n^0)  
\end{equation}
The residual $\Delta W_n$ is formulated as:
\begin{equation}\label{eq:3}
    \Delta W_n = P_n^T \otimes Q_n
\end{equation}
where $\otimes$ means matrix multiplication. $P_n\in\mathbb{R}^{L \times C_{out}}$ and $ Q_n\in\mathbb{R}^{L \times C_{in}}$ are two sets of learnable tokens, and $L \ll C_{out} \leq C_{in}$ is a hyper-parameter. Note that $\Delta W_n$ is a low-rank matrix, since its rank is at most $L$. By decomposing $\Delta W_n$ into $P_n$ and $Q_n$, the number of parameters is reduced to $L \times (C_{out}+C_{in})$.

\subsubsection{Learnable Scaling Factors}
To adaptively control the amount of residual $\Delta W_n$, we apply learnable scaling factors for each token in $P_n$ and $Q_n$ as:
\begin{equation}
\begin{aligned}
    P_n &= A \odot P_n, \quad A = (\alpha_1, \alpha_2, ..., \alpha_L)\\
    Q_n &= B \odot Q_n, \quad B = (\beta_1, \beta_2, ..., \beta_L)
\end{aligned}
\end{equation}
where $A, B\in \mathbb{R}^{L\times 1}$ are scaling vectors, and $\odot$ indicates element-wise product. Note that $A, B$ are simply expanded to the corresponding dimensions before multiplication. The factors $\alpha_l, \beta_l$ in $A, B$, initialized with small values, are to scale the $l$-th token in $P_n, Q_n$.

\begin{figure*}[t]
\centering
\includegraphics[width=1\textwidth]{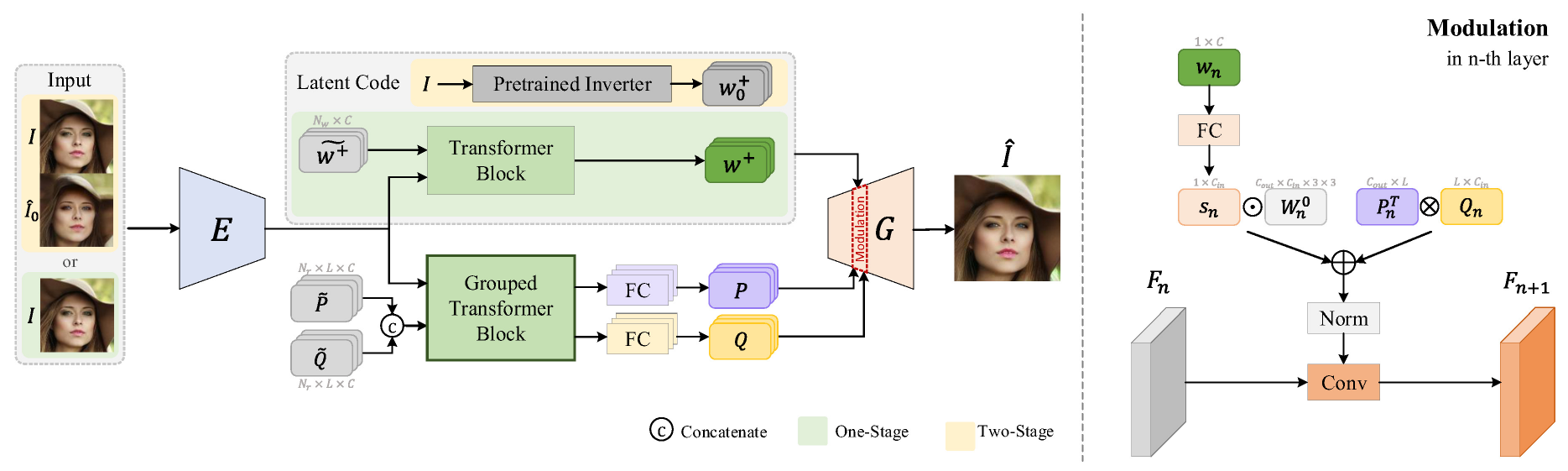}
\caption{The RefineStyle applied in image inversion. Given a real image $I$, the model specifies $w^+$ code to form dynamic convolution kernels, and tokens $P$ and $Q$ to refine them, as shown on the right. Two training strategies are distinguished with different background color on the left. In one-stage training(green), initial $\Tilde{w^+}$, $\Tilde{P}$ and $\Tilde{Q}$ are updated with real image features encoded by $E$. In two-stage training(grey), a pre-trained inverter gives $w_0^+$ and initial inversion $\hat{I}_0$. The model takes the concatenated $[I,\hat{I}_0]$ as input to update $\Tilde{P}$ and $\Tilde{Q}$, and fix $w_0^+$ for modulation.}
\label{fig:fig1}
\end{figure*}

\subsection{Application on Image Inversion}
\subsubsection{Grouped Transformer Blocks}
To verify the refining scheme, we apply it in image inversion with an efficient model built on grouped transformer blocks to produce the token sets $P$ and $Q$, as shown in Fig. \ref{fig:fig1}. These blocks incorporate grouped self-attention and cross-attention. We assume $\Tilde{P}, \Tilde{Q}\in \mathbb{R}^{N_r \times L \times C}$ as initial trainable tokens, where $N_r$ is the number of layers to refine and $C$ is set to 512. The grouped self-attention divides tokens into $N_r$ groups, and performs interaction among $L$ tokens within the same group, which is illustrated as:
\begin{equation}
\begin{aligned}
  GroupedAttn(q_i,k_i,v_i)= &\text{Softmax}(\frac{q_i {k_i}^T}{\sqrt{C}}) v_i\\
  q_i,k_i,v_i\in\mathbb{R}^{L \times C}, \quad &i=0,1,...,N_r
\end{aligned}
\end{equation}
where $q_i$, $k_i$ and $v_i$ represent the query, key and value mapped from tokens in the $i$-th group. Post self-attention, all groups of tokens are considered as a whole. Then in the cross-attention, queries projected from these tokens interact with keys and values from image features of different resolutions, given by a feature pyramid encoder $E$. To fulfill the dimension requirement in Eq.(\ref{eq:3}), we take the $n$-th group of tokens from $\Tilde{P}, \Tilde{Q}$ as $P_n, Q_n$ for the $n$-th layer and utilize two projection heads to change the channel dimensions into $C_{out}, C_{in}$. 
\subsubsection{Training Strategies}
There are two training strategies for our inversion model, one- and two-stage, distinguished by background colors in Fig. \ref{fig:fig1}. The former trains the model to invert images by $w^+$ and dynamic residuals $\Delta W$ jointly, while the latter takes the initial inverted images and $w_0^+$ as input and only optimizes the model for $\Delta W$. 

\textbf{The one-stage model} shares the image encoder $E$ for $w^+$ and $\Delta W$. Following~\cite{hu2022style}, we use the initial style codes $\Tilde{w^+}$ as queries, which are mapped from Gaussian noise by the MLP in pre-trained StyleGAN, and regard the multi-resolution image features from $E$ as keys and values. Three transformer blocks, consisting of both self- and cross- attention, sequentially update queries into the consequent code $w^+$. Additional grouped transformer blocks are employed to update the initial $\tilde{P}$ and $\tilde{Q}$ by image features, and form $\Delta W$ through projection heads and multiplication.

\textbf{The two-stage model} begins with the initial inverted image $\hat{I}_0$ and code $w_0^+$ from a pre-trained inversion model~\cite{hu2022style}. It concatenates the real image $I$ and $\hat{I_0}$ along the channel dimension, resulting in 6-channel input for encoder $E$ to generate multi-resolution features. Three grouped transformer blocks then use these features to acquire $\Delta W$. Note that the pre-trained inverter is frozen.

\textbf{Losses.} During training, the generator $G$ is strictly fixed. The MLP mapping $z$ to $w$ is frozen in the two-stage model and trainable in one-stage. Both models are trained by the following losses. The $L_2$ loss between the real image $I$ and the inverted image $\hat{I}$ is calculated as: $L_2=\Vert I-\hat{I}\Vert_2$.
We also adopt the LPIPS~\cite{zhang2018unreasonable} loss $L_{LPIPS}=\Vert F(I)-F(\hat{I}) \Vert_2$, based on Inception network $F(\cdot)$. 
Moreover, the ID loss $L_{ID}=1-\langle R(I),R(\hat{I}) \rangle$ is applied to preserve the identity, where $\langle\cdot,\cdot\rangle$ denotes the cosine similarity and $R(\cdot)$ is a pre-trained ArcFace model~\cite{deng2019arcface}.

The full objective is expressed as follows,
\begin{equation}
\label{eq:eq_rec}
    L_{rec}=\lambda_{2} L_{2} + \lambda_{lpips} L_{LPIPS} + \lambda_{id} L_{ID}
\end{equation}
where $\lambda_{2}$, $\lambda_{lpips}$ and $\lambda_{id}$ are hyper-parameters for each term.

\begin{figure*}[!bthp]
\centering
\includegraphics[width=1\textwidth]{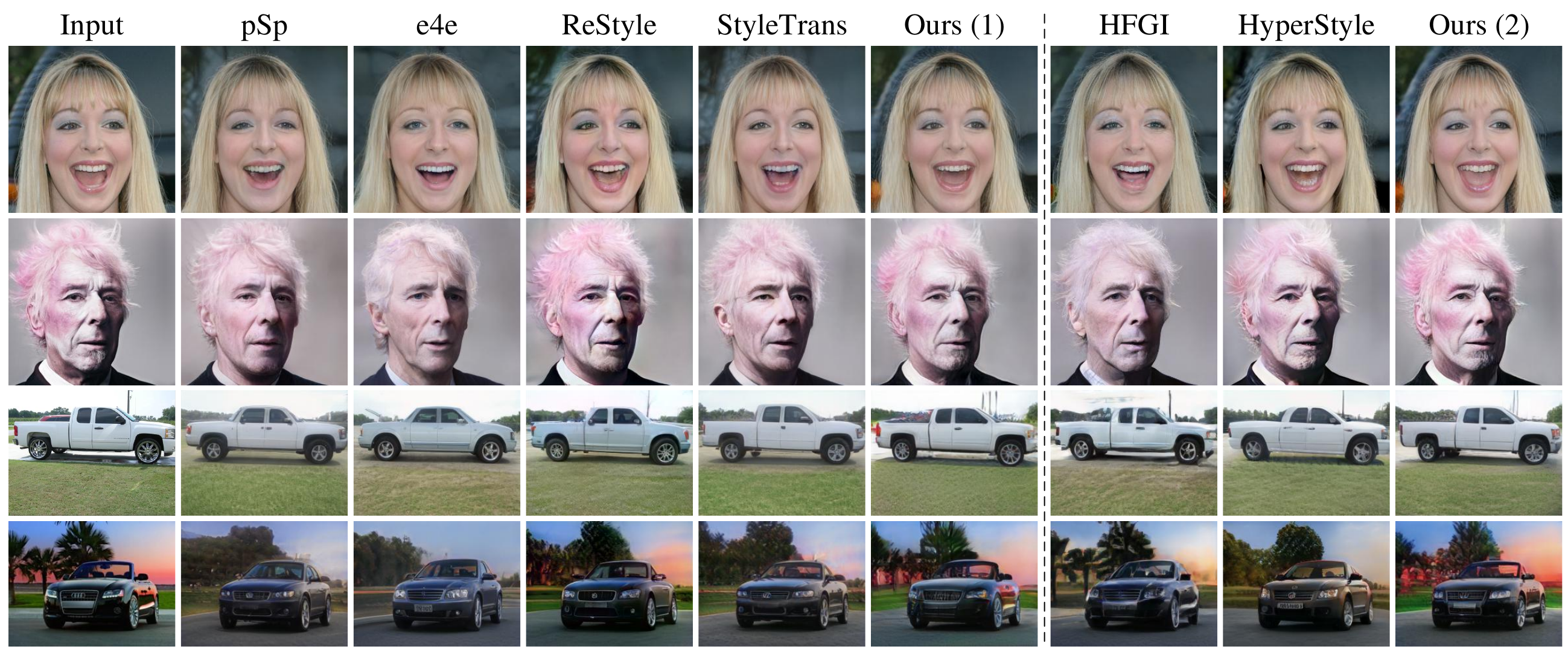}
\caption{Qualitative comparison of image inversion. \textit{Ours(1)} and \textit{Ours(2)} denote the one- and two-stage models. RefineStyle excels in reconstructing details, such as mouth in the 1-st row, hair in the 2-nd row and background in the 4-th row.}
\label{fig:inversion}
\end{figure*}

\subsection{Application on Domain Adaptation}
The refining scheme can be employed for domain adaptation to turn the pre-trained $G$ into $\hat{G}$ at target domain. $P$ and $Q$ are expected to account for style change, so we choose to optimize them directly.
Given source and target text descriptions $t_{src}$ and $t_{tar}$, we employ direction loss $L_{dir}$ as in~\cite{gal2021stylegan}, which utilizes CLIP image encoder $E_I$ and text encoder $E_T$ to ensure the desired image offset $\Delta I=E_I(\hat{G}(w^+))-E_I(G(w^+))$ consistent with text offset $\Delta T=E_T(t_{tar})-E_T(t_{src})$. Formally, we calculate $L_{dir}$ as follows:
\begin{equation}\label{eq:eq_nada}
    L_{dir}=1-\frac{\Delta I}{|\Delta I|}\frac{\Delta T}{|\Delta T|}
\end{equation}

The target style can also be specified by reference images. We investigate the extreme case that only one image $I_r$ is available, and apply the same losses in~\cite{Zhang2022GeneralizedOD} to train $P$ and $Q$. Specifically, one loss term is to reconstruct $I_r$ with $\hat{G}$, formulated as $L_{rec}$ in Eq.(\ref{eq:eq_rec}). Another term $L_{sty}$ employs sliced Wasserstein distance (SWD)~\cite{bonneel2015sliced} to ensure random synthesis $\hat{I}_g$ mimic the style of $I_r$. $\hat{I}_g$ is generated with a concatenation code $w_g=[w_c,w_r]\in\mathcal{W}^+$, where content code $w_c$ is mapped from random $z$, and style code $w_r$ is from inversion of $I_r$. Since the later stage code determines the texture details, the synthesis $\hat{I}_g$ takes the general style of $I_r$, facilitating more stable training at the beginning.

\begin{figure*}[!tbhp]
\centering
\includegraphics[width=1\textwidth]{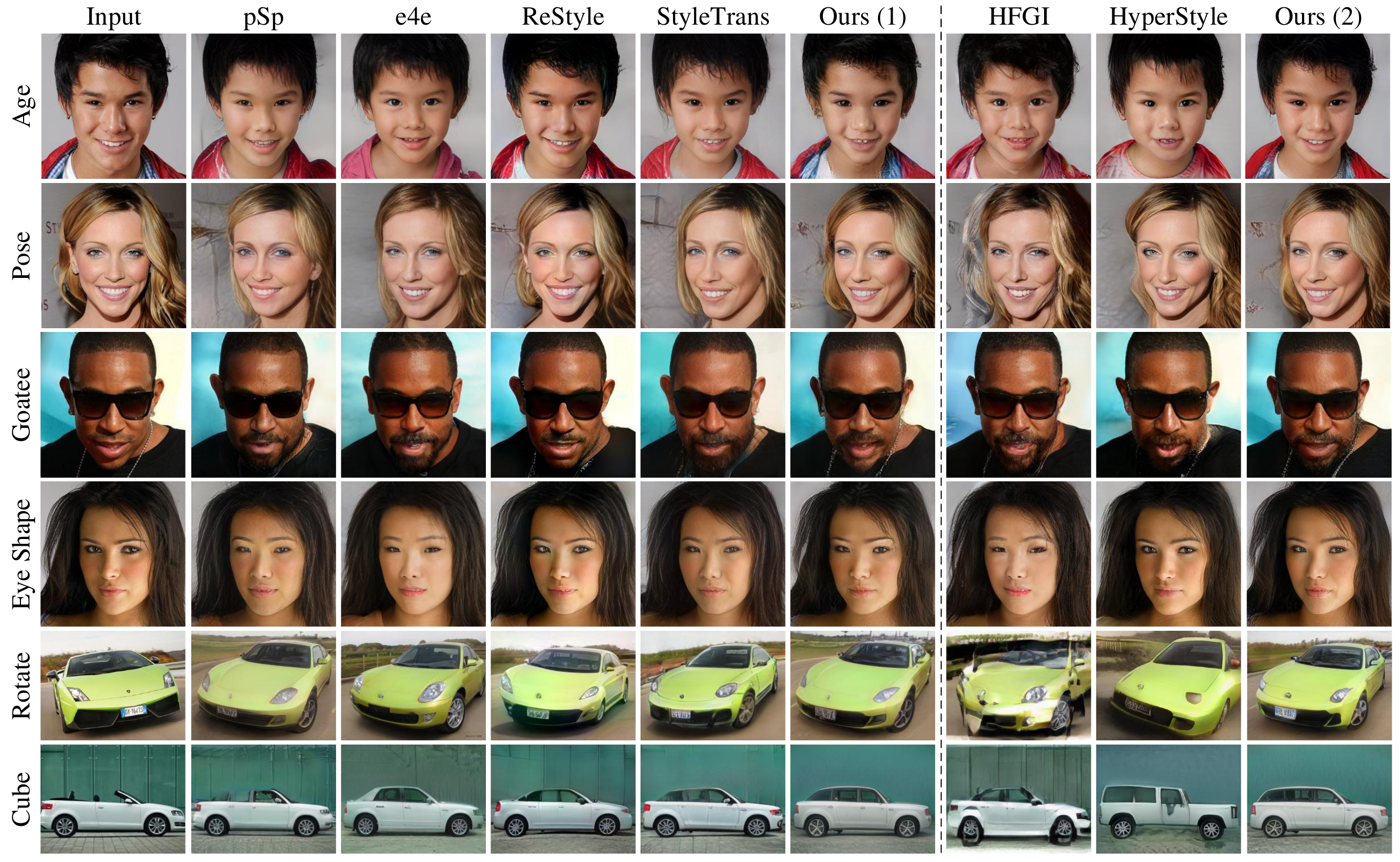}
\caption{Qualitative comparison of label-based editing. \textit{Ours(1)} and \textit{Ours(2)} denote the one- and two-stage models. RefineStyle preserves details of other attributes while editing specific one, such as clothing in the 1-st row and the necklace in the 3-rd row. }
\label{fig:labeledit}
\end{figure*}

\section{Experiments}
\subsection{Implementation Details}
All experiments are implemented on StyleGAN2~\cite{karras2020analyzing} pre-trained on FFHQ~\cite{karras2019style} and LSUN Cars~\cite{yu2015lsun} datasets. In the inversion task, we train models on FFHQ dataset and evaluated on CelebA-HQ~\cite{karras2017progressive} test set for face domain, and use Stanford Cars~\cite{krause20133d} dataset for car domain's training and evaluation. For two-stage model, we use pre-trained Style Transformer~\cite{hu2022style} to provide style codes and initial inverted images. In the domain adaptation task, we freeze the pre-trained FFHQ StyleGAN2 and train the token sets following~\cite{gal2021stylegan,Zhang2022GeneralizedOD}. In one-shot domain adaption, our inversion model can obtain the reference image's token sets for better initialization. More details are given in the Supplementary Material.

\subsection{Inversion Results}
We compare our one-stage model with pSp~\cite{richardson2021encoding}, e4e~\cite{tov2021designing}, ReStyle~\cite{alaluf2021restyle}, Style Transformer~\cite{hu2022style}, and our two-stage model with HFGI~\cite{wang2022high} and HyperStyle~\cite{alaluf2022hyperstyle}. Quantitative results are illustrated in Table~\ref{tab:inversion}. The inversion distortion is evaluated by MSE, LPIPS and identity similarity (ID), the last only for face domain. We also report number of parameters, FLOPs and inference time. Our one- and two-stage models both achieve the lowest distortion with limited inference time and lightweight design. 
As Fig. \ref{fig:inversion} shows, RefineStyle reconstructs the out-of-domain regions faithfully, such as open mouth and exaggerated hairstyle.

\begin{table*}[!hbtp]
\renewcommand\arraystretch{1.1}
\caption{Quantitative comparison of image inversion on CelebA-HQ and Stanford Cars datasets.}
\label{tab:inversion}
\centering
\resizebox{1\textwidth}{!}{
\begin{tabular}{l|c|c|c|c|c|c|c|c|c|c|c|c}
\hline
\multirow{3}{*}{\textbf{Methods}} & \multicolumn{8}{c|}{\textbf{Face}}&\multicolumn{4}{c}{\textbf{Cars}}\\ 
\cline{2-9}\cline{10-13}
& \multicolumn{6}{c|}{\textbf{Inversion}}& \multicolumn{2}{c|}{\textbf{Editing}}&\multicolumn{2}{c|}{\textbf{Inversion}}& \multicolumn{2}{c}{\textbf{Editing}}\\ \cline{2-9}\cline{10-13}
                 & MSE$\downarrow$ & LPIPS$\downarrow$ & ID$\uparrow$ & Params$\downarrow$ & FLOPs$\downarrow$ & Time$\downarrow$ & ID$\uparrow$ & User$\uparrow$ & MSE$\downarrow$  & LPIPS$\downarrow$ & FID$\downarrow$ & User$\uparrow$ \\ \hline
  pSp            & 0.037          & 0.169          & 0.56         & 267M          & 72.5G         & 0.067s         & 0.44         & 10.3\%         & 0.115          & 0.298 & 39.6         & 19.0\%         \\
  e4e            & 0.050          & 0.209          & 0.50         & 267M          & 72.5G         & 0.066s         & 0.40         & 13.8\%         & 0.110          & 0.314 & 31.7         & 13.8\%         \\
  ReStyle        & 0.031          & 0.137          & 0.66         & 205M          & 183.1G        & 0.273s         & 0.59         & 25.9\%         & 0.075          & 0.225 & 27.7         & 8.6\%         \\
  StyleTrans     & 0.036          & 0.166          & 0.59         & \textbf{40.6M}& \textbf{36.3G}& \textbf{0.044s}& 0.46         & 15.5\%         & 0.089          & 0.245 & 36.4         & 17.2\%         \\
  Ours (1-stage) & \textbf{0.017} & \textbf{0.073} & \textbf{0.75}& 46.4M         & 43.0G         & 0.045s         & \textbf{0.60} & \textbf{34.5\%} & \textbf{0.067} & \textbf{0.173} & \textbf{27.1} & \textbf{41.4\%} \\
  \hline
  HFGI           & 0.025          & 0.127          & 0.67         & \textbf{6.2M} & \textbf{29.8G}& 0.097s         & 0.48         & 12.1\%         & \textbf{0.042} & 0.194 & 43.9         & 5.2\%         \\
  HyperStyle     & 0.019          & 0.093          & 0.76         & 332M          & 166.6G        & 0.910s         & 0.60         & 36.2\%         & 0.074          & 0.291 & 25.7         & 27.6\%         \\   
  Ours (2-stage) & \textbf{0.019} & \textbf{0.082} & \textbf{0.76}& 47.1M         & 51.1G         & \textbf{0.046s}& \textbf{0.61} & \textbf{51.7\%}  & 0.064          & \textbf{0.185} & \textbf{23.8} & \textbf{67.2\% } \\
  \hline
\end{tabular}
}
\end{table*}

\subsection{Editing Results}
Label-based editing alters the latent code along a label-specific semantic direction, which can assess the editability of inversion methods. We adopt InterFaceGAN~\cite{shen2020interpreting} and StyleSpace~\cite{wu2021stylespace} to edit style codes $w^+\in\mathcal{W}^+$ for face domain, and GANSpace~\cite{harkonen2020ganspace} in $\mathcal{S}$ space for car domain. Then we complement the edited codes with our proposed residuals. Quantitative results are shown in Table~\ref{tab:inversion}. For face domain, we calculate ID between edited images and real images, and report the average scores among 5 attributes. For car domain, we assess averaged FID among 3 attributes. We further conduct user studies in both domains. We ask 58 volunteers to evaluate the visual quality and editing conformity of edited images. Qualitative results are provided in Fig. \ref{fig:labeledit}, with the top two rows showing edits from InterFaceGAN, the 3-rd and 4-th from StyleSpace, and the last two from GANSpace. As shown, the residuals effectively preserve the content and prevent unwanted changes.

\begin{figure*}[!b]
\centering
\includegraphics[width=1\textwidth]{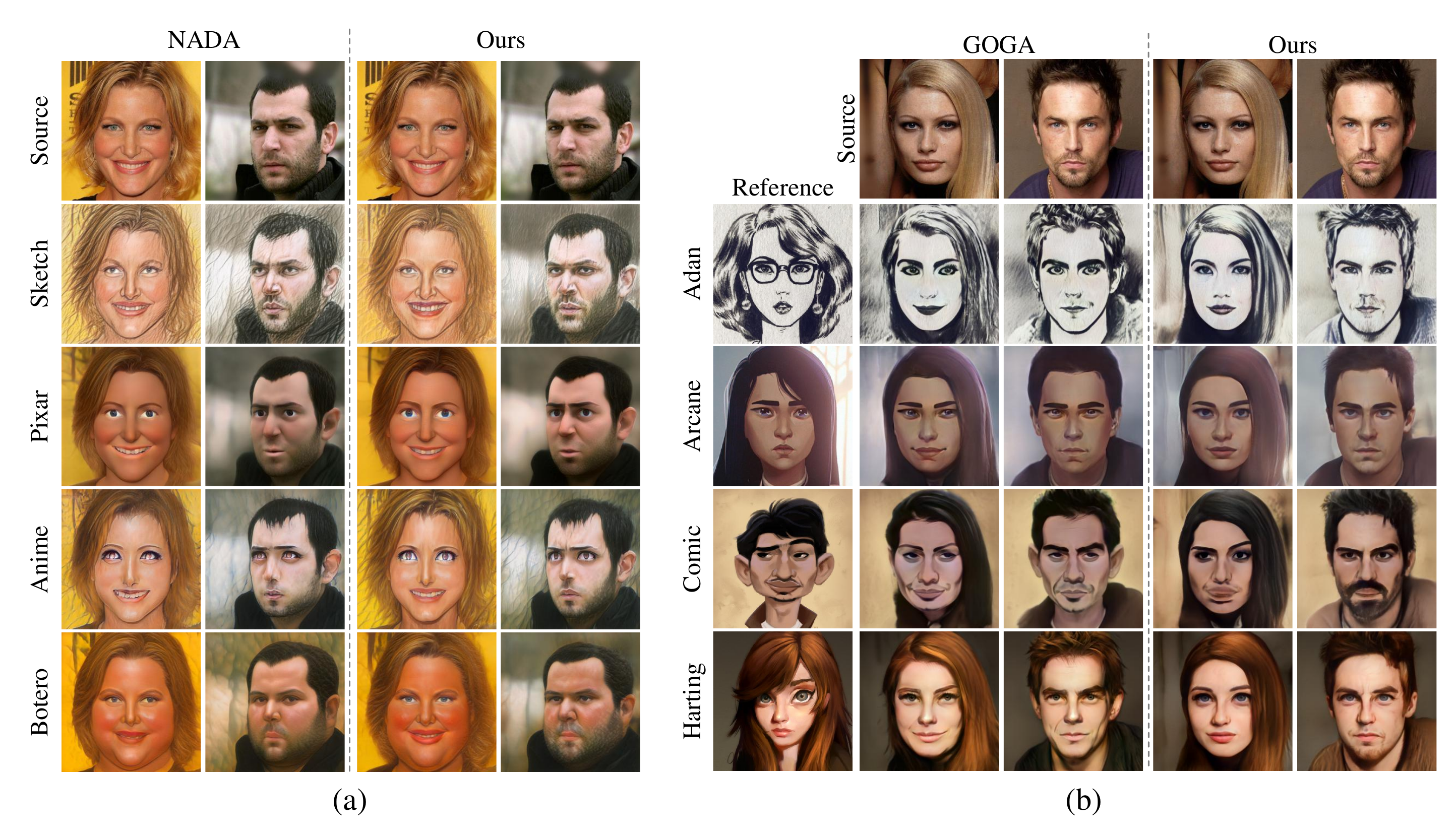}
\caption{(a) Text-driven domain adaptation synthesis compared with StyleGAN-NADA~\cite{gal2021stylegan}. (b) One-shot domain adaptation synthesis compared with GOGA~\cite{Zhang2022GeneralizedOD}}.
\label{fig:adapt}
\end{figure*}

\subsection{Domain Adaptation Results}
We train our domain adaptation models following the strategies of StyleGAN-NADA~\cite{gal2021stylegan} and GOGA~\cite{Zhang2022GeneralizedOD}. Qualitative and quantitative results compared with StyleGAN-NADA are shown in Fig. \ref{fig:adapt}(a) and Table~\ref{tab:adapt}. We evaluate the methods by ID and text consistency (TC), which calculates the cosine similarity between the CLIP embeddings of target text and transferred images. The scores are averaged over 2000 generated images. We also list the trainable parameters. Our method achieves favourable performance with an efficient and lightweight model. Comparisons with GOGA can be seen in Fig. \ref{fig:adapt}(b) and Table~\ref{tab:oneshot}. We report ID, Normalized Mean Error (NME) and number of parameters. NME measures the error of facial landmarks between real and adapted images. Results show that our model has the capability to transfer styles using single reference image while keeping the identity more faithfully.

\begin{table}[!ht]
\caption{Quantitative comparison of text-driven domain adaptation with StyleGAN-NADA.}
\label{tab:adapt}
\centering
\setlength{\tabcolsep}{4pt}
\begin{tabular}{l|c|c|c|c|c|c|c|c|c}
\hline
\multirow{2}{*}{Methods} & \multicolumn{2}{c|}{Sketch} & \multicolumn{2}{c|}{Pixar} & \multicolumn{2}{c|}{Anime} & \multicolumn{2}{c|}{Botero} & \multirow{2}{*}{Params$\downarrow$}\\ 
\cline{2-9}
             & ID$\uparrow$ & TC$\uparrow$ & ID$\uparrow$ & TC$\uparrow$ & ID$\uparrow$ & TC$\uparrow$ & ID$\uparrow$ & TC$\uparrow$ \\ \hline
NADA         & 0.47          & 0.25          & \textbf{0.39} & 0.31          & 0.27          & 0.24         & 0.27          & 0.33          & 26.7M \\
Ours         & \textbf{0.51} & \textbf{0.47} & 0.38          & \textbf{0.62} & \textbf{0.28} & \textbf{0.53}& \textbf{0.34} & \textbf{0.42} & \textbf{2.98M} \\ \hline
\end{tabular}
\end{table}

\begin{table}[!ht]
\caption{Quantitative comparison of one-shot domain adaptation with GOGA..}
\label{tab:oneshot}
\centering
\setlength{\tabcolsep}{4pt}
\begin{tabular}{l|c|c|c|c|c|c|c|c|c}
\hline
\multirow{2}{*}{Methods} & \multicolumn{2}{c|}{Adan} & \multicolumn{2}{c|}{Arcane} & \multicolumn{2}{c|}{Comic} & \multicolumn{2}{c|}{Harting} & \multirow{2}{*}{Params$\downarrow$}\\ 
\cline{2-9} 
             & ID$\uparrow$ & NME$\downarrow$ & ID$\uparrow$ & NME$\downarrow$ & ID$\uparrow$ & NME$\downarrow$ & ID$\uparrow$ & NME$\downarrow$ \\ \hline
GOGA         & 0.11          & 0.026          & 0.12          & 0.023          & 0.11          & 0.029         & 0.10          & 0.026          & 28.3M\\
Ours         & \textbf{0.20} & \textbf{0.022} & \textbf{0.19} & \textbf{0.021} & \textbf{0.17} & \textbf{0.027}& \textbf{0.17} & \textbf{0.025} & \textbf{2.98M} \\ \hline
\end{tabular}
\end{table}

\subsection{Ablation Studies}
To further investigate the effectiveness of structures and training strategies of RefineStyle inversion model, we conduct ablation experiments on CelebA-HQ test set. $N_r$ is the number of layers to refine, $L$ is the number of tokens for each layer, \textit{group} denotes grouped self-attention, and \textit{learnable factor} means learnable scaling factors for token sets. As shown in Table~\ref{tab:ablation}, the optimal $N_r$ is set to 10 for one-stage training and 17 for two-stage, indicating that either the last 10 convolution layers of generator are refined or all layers are. Note that we do not refine the toRGB layers. The grouped self-attention and learnable scaling factors prove valid in both methods, as does increasing $L$.

To validate the versatility of the low-rank residuals, we further conduct an ablation study on different pre-trained inverters in two-stage RefineStyle inversion model. We compare our models with HyperStyle, which originally uses pre-trained e4e model, and is retrained based on pre-trained StyleTrans, shown in the 2-nd and 3-rd rows in Table~\ref{tab:ablation_hyper}. Our two-stage RefineStyle inversion models using pSp, e4e and StyleTrans as the pre-trained inverters, all achieve comparable MSE and ID metrics as HyperStyle, and excel in the LPIPS metric, demonstrating the effectiveness of the proposed low-rank residuals.

\begin{table}[!htb]
\caption{Ablation study on the structures and training strategies of RefineStyle inversion models}
\label{tab:ablation}
\renewcommand\arraystretch{1.1}
\centering
\setlength{\tabcolsep}{4pt}
\begin{tabular}{l|c|c|c|c}
\hline
  Methods         & $N_r$  & $L$  & MSE $\downarrow$  & LPIPS $\downarrow$\\ \hline
  \multirow{2}{*}{w/o group, learnable factor} & 10  & 8  & 0.034 & 0.157\\
                                               & 10  & 32 & 0.023 & 0.101\\\hdashline
  \multirow{2}{*}{w/o learnable factor}        & 10  & 32 & 0.022 & 0.098\\
                                               & 17  & 32 & 0.020 & 0.084\\\hdashline
  \textbf{Ours (2-stage)}                      & \textbf{17}  & \textbf{32} & \textbf{0.019}  & \textbf{0.082}\\\hline
  \multirow{2}{*}{w/o learnable factor}        & 10  & 32 & 0.022 & 0.102\\
                                               & 17  & 32 & 0.027 & 0.123\\\hdashline
  \textbf{Ours (1-stage)}                      & \textbf{10}  & \textbf{32} & \textbf{0.017}  & \textbf{0.073}\\ \hline
\end{tabular}
\end{table}

\begin{table}[!htb]
\caption{Ablation study on the pre-trained inverters in two-stage training.}
\label{tab:ablation_hyper}
\renewcommand\arraystretch{1.1}
\centering
\setlength{\tabcolsep}{4pt}
\begin{tabular}{l|c|c|c}
\hline
  Methods                 & MSE $\downarrow$ & LPIPS $\downarrow$ & ID $\uparrow$\\ \hline
  HyperStyle (e4e)        & 0.019            & 0.093              & 0.76 \\
  HyperStyle (StyleTrans) & 0.020            & 0.095              & 0.75 \\
  Ours (pSp)              & 0.019            & 0.087              & 0.76 \\
  Ours (e4e)              & 0.019            & 0.084              & 0.76 \\
  \textbf{Ours (StyleTrans)}& \textbf{0.019}   & \textbf{0.082}     & \textbf{0.76} \\\hline
\end{tabular}
\end{table}

\section{Discussion}
Extensive experiments demonstrate the effectiveness of our proposed low-rank residuals for StyleGAN inversion and domain adaptation. As in LoRA~\cite{hu2021lora}, low-rank residuals are also applied to large language model and diffusion text-image models, adapting the static weights in transformers. Different from LoRA, our residuals refine dynamic convolution kernels, which are modulated by style code $w^+\in \mathcal{W}^+$. Since $w^+$ is assigned by single image, specific image features are embed into the dynamic kernels. Particularly, our inversion model applies an encoder to generate distinctive low-rank residuals for each input image.

\section{Conclusion}
This paper proposes to refine dynamic convolution kernels in pre-trained StyleGAN, enabling high-quality out-of-domain synthesis. In addition to the original dynamic modulation in StyleGAN, our model further refines kernels according to an input image or specific domain guidance. To maintain simplicity post-refinement, we learn two token sets for each layer of generator, and multiply them to form low-rank residuals. Our scheme can be utilized for both image inversion and domain adaptation. For image inversion, we design grouped transformer blocks with one- and two-stage training strategies. For domain adaptation, we show that the learnable token sets can be directed towards target domain. Experiments demonstrate our scheme's superiority and broad application potential.

\textbf{Acknowledgements.}
This work is supported by the Science and Technology Commission of Shanghai Municipality under Grant No. 22511105800, 19511120800 and 22DZ2229004.
%
%
\FloatBarrier

\begin{thebibliography}{10}
\providecommand{\url}[1]{\texttt{#1}}
\providecommand{\urlprefix}{URL }
\providecommand{\doi}[1]{https://doi.org/#1}

\bibitem{abdal2019image2stylegan}
Abdal, R., Qin, Y., Wonka, P.: Image2stylegan: How to embed images into the stylegan latent space? In: Proceedings of the IEEE/CVF International Conference on Computer Vision. pp. 4432--4441 (2019)

\bibitem{abdal2020image2stylegan++}
Abdal, R., Qin, Y., Wonka, P.: Image2stylegan++: How to edit the embedded images? In: Proceedings of the IEEE/CVF conference on computer vision and pattern recognition. pp. 8296--8305 (2020)

\bibitem{alaluf2021restyle}
Alaluf, Y., Patashnik, O., Cohen-Or, D.: Restyle: A residual-based stylegan encoder via iterative refinement. In: Proceedings of the IEEE/CVF International Conference on Computer Vision. pp. 6711--6720 (2021)

\bibitem{alaluf2022hyperstyle}
Alaluf, Y., Tov, O., Mokady, R., Gal, R., Bermano, A.: Hyperstyle: Stylegan inversion with hypernetworks for real image editing. In: Proceedings of the IEEE/CVF Conference on Computer Vision and Pattern Recognition. pp. 18511--18521 (2022)

\bibitem{bonneel2015sliced}
Bonneel, N., Rabin, J., Peyr{\'e}, G., Pfister, H.: Sliced and radon wasserstein barycenters of measures. Journal of Mathematical Imaging and Vision  \textbf{51}(1),  22--45 (2015)

\bibitem{deng2019arcface}
Deng, J., Guo, J., Xue, N., Zafeiriou, S.: Arcface: Additive angular margin loss for deep face recognition. In: Proceedings of the IEEE/CVF Conference on Computer Vision and Pattern Recognition. pp. 4690--4699 (2019)

\bibitem{dinh2022hyperinverter}
Dinh, T.M., Tran, A.T., Nguyen, R., Hua, B.S.: Hyperinverter: Improving stylegan inversion via hypernetwork. In: Proceedings of the IEEE/CVF Conference on Computer Vision and Pattern Recognition. pp. 11389--11398 (2022)

\bibitem{gal2021stylegan}
Gal, R., Patashnik, O., Maron, H., Chechik, G., Cohen-Or, D.: Stylegan-nada: Clip-guided domain adaptation of image generators. arXiv preprint arXiv:2108.00946  (2021)

\bibitem{harkonen2020ganspace}
H{\"a}rk{\"o}nen, E., Hertzmann, A., Lehtinen, J., Paris, S.: Ganspace: Discovering interpretable gan controls. Advances in Neural Information Processing Systems  \textbf{33},  9841--9850 (2020)

\bibitem{hu2021lora}
Hu, E.J., Shen, Y., Wallis, P., Allen-Zhu, Z., Li, Y., Wang, S., Wang, L., Chen, W.: Lora: Low-rank adaptation of large language models. arXiv preprint arXiv:2106.09685  (2021)

\bibitem{hu2022style}
Hu, X., Huang, Q., Shi, Z., Li, S., Gao, C., Sun, L., Li, Q.: Style transformer for image inversion and editing. In: Proceedings of the IEEE/CVF Conference on Computer Vision and Pattern Recognition. pp. 11337--11346 (2022)

\bibitem{karras2017progressive}
Karras, T., Aila, T., Laine, S., Lehtinen, J.: Progressive growing of gans for improved quality, stability, and variation. arXiv preprint arXiv:1710.10196  (2017)

\bibitem{karras2021alias}
Karras, T., Aittala, M., Laine, S., H{\"a}rk{\"o}nen, E., Hellsten, J., Lehtinen, J., Aila, T.: Alias-free generative adversarial networks. Advances in Neural Information Processing Systems  \textbf{34},  852--863 (2021)

\bibitem{karras2019style}
Karras, T., Laine, S., Aila, T.: A style-based generator architecture for generative adversarial networks. In: Proceedings of the IEEE/CVF Conference on Computer Vision and Pattern Recognition. pp. 4401--4410 (2019)

\bibitem{karras2020analyzing}
Karras, T., Laine, S., Aittala, M., Hellsten, J., Lehtinen, J., Aila, T.: Analyzing and improving the image quality of stylegan. In: Proceedings of the IEEE/CVF Conference on Computer Vision and Pattern Recognition. pp. 8110--8119 (2020)

\bibitem{krause20133d}
Krause, J., Stark, M., Deng, J., Fei-Fei, L.: 3d object representations for fine-grained categorization. In: Proceedings of the IEEE international conference on computer vision workshops. pp. 554--561 (2013)

\bibitem{liu2023delving}
Liu, H., Song, Y., Chen, Q.: Delving stylegan inversion for image editing: A foundation latent space viewpoint. In: Proceedings of the IEEE/CVF conference on computer vision and pattern recognition. pp. 10072--10082 (2023)

\bibitem{ojha2021few}
Ojha, U., Li, Y., Lu, J., Efros, A.A., Lee, Y.J., Shechtman, E., Zhang, R.: Few-shot image generation via cross-domain correspondence. In: Proceedings of the IEEE/CVF Conference on Computer Vision and Pattern Recognition. pp. 10743--10752 (2021)

\bibitem{pehlivan2023styleres}
Pehlivan, H., Dalva, Y., Dundar, A.: Styleres: Transforming the residuals for real image editing with stylegan. In: Proceedings of the IEEE/CVF Conference on Computer Vision and Pattern Recognition. pp. 1828--1837 (2023)

\bibitem{richardson2021encoding}
Richardson, E., Alaluf, Y., Patashnik, O., Nitzan, Y., Azar, Y., Shapiro, S., Cohen-Or, D.: Encoding in style: a stylegan encoder for image-to-image translation. In: Proceedings of the IEEE/CVF Conference on Computer Vision and Pattern Recognition. pp. 2287--2296 (2021)

\bibitem{roich2021pivotal}
Roich, D., Mokady, R., Bermano, A.H., Cohen-Or, D.: Pivotal tuning for latent-based editing of real images. arXiv preprint arXiv:2106.05744  (2021)

\bibitem{shen2020interpreting}
Shen, Y., Gu, J., Tang, X., Zhou, B.: Interpreting the latent space of gans for semantic face editing. In: Proceedings of the IEEE/CVF conference on computer vision and pattern recognition. pp. 9243--9252 (2020)

\bibitem{tov2021designing}
Tov, O., Alaluf, Y., Nitzan, Y., Patashnik, O., Cohen-Or, D.: Designing an encoder for stylegan image manipulation. ACM Transactions on Graphics (TOG)  \textbf{40}(4),  1--14 (2021)

\bibitem{wang2022high}
Wang, T., Zhang, Y., Fan, Y., Wang, J., Chen, Q.: High-fidelity gan inversion for image attribute editing. In: Proceedings of the IEEE/CVF Conference on Computer Vision and Pattern Recognition. pp. 11379--11388 (2022)

\bibitem{wang2018transferring}
Wang, Y., Wu, C., Herranz, L., van~de Weijer, J., Gonzalez-Garcia, A., Raducanu, B.: Transferring gans: generating images from limited data. In: Proceedings of the European Conference on Computer Vision (ECCV). pp. 218--234 (2018)

\bibitem{wu2021stylespace}
Wu, Z., Lischinski, D., Shechtman, E.: Stylespace analysis: Disentangled controls for stylegan image generation. In: Proceedings of the IEEE/CVF Conference on Computer Vision and Pattern Recognition. pp. 12863--12872 (2021)

\bibitem{yang2021one}
Yang, C., Shen, Y., Zhang, Z., Xu, Y., Zhu, J., Wu, Z., Zhou, B.: One-shot generative domain adaptation. arXiv preprint arXiv:2111.09876  (2021)

\bibitem{yang2022Pastiche}
Yang, S., Jiang, L., Liu, Z., Loy, C.C.: Pastiche master: Exemplar-based high-resolution portrait style transfer. In: CVPR (2022)

\bibitem{yu2015lsun}
Yu, F., Seff, A., Zhang, Y., Song, S., Funkhouser, T., Xiao, J.: Lsun: Construction of a large-scale image dataset using deep learning with humans in the loop. arXiv preprint arXiv:1506.03365  (2015)

\bibitem{zhang2018unreasonable}
Zhang, R., Isola, P., Efros, A.A., Shechtman, E., Wang, O.: The unreasonable effectiveness of deep features as a perceptual metric. In: Proceedings of the IEEE conference on computer vision and pattern recognition. pp. 586--595 (2018)

\bibitem{Zhang2022GeneralizedOD}
Zhang, Z., Liu, Y., Han, C., Guo, T., Yao, T., Mei, T.: Generalized one-shot domain adaptation of generative adversarial networks. In: Advances in Neural Information Processing Systems (2022)

\bibitem{zhu2016generative}
Zhu, J.Y., Kr{\"a}henb{\"u}hl, P., Shechtman, E., Efros, A.A.: Generative visual manipulation on the natural image manifold. In: European conference on computer vision. pp. 597--613. Springer (2016)

\end{thebibliography}
%

\appendix
\section{Experimental Settings}\label{secA1}
\subsection{Inversion models}
We adopt pre-trained StyleGAN2 generator in all experiments. The parameters of backbone are fixed during training, while the mapping network is trainable in our 
one-stage model. 
We apply pre-trained Arcface model as encoder $E$, and randomly re-initialize the input layer for the 6-channel input in our two-stage training. Following the design of the transformer blocks in StyleTrans, multi-head attention is applied in our grouped transformer blocks.
The number of heads is set to 4, and the dimension of each head is 512. We train the models using Ranger optimizer, which is a combination of Rectified Adam and the Lookahead technique. 
Our one-stage model is trained for $1 \times 10^{6}$ iterations with a batch size of 8, and a learning rate of $1 \times 10^{-4}$. Our two-stage model is trained for $7 \times 10^{5}$ iterations with a batch size of 8 and learning rate of 
$1 \times 10^{-3}$. All experiments are implemented on 2 NVIDIA RTX 2080 GPUs.
\subsection{Domain adaptation models}
\begin{figure}[H]
\centering
\includegraphics[width=0.7\textwidth]{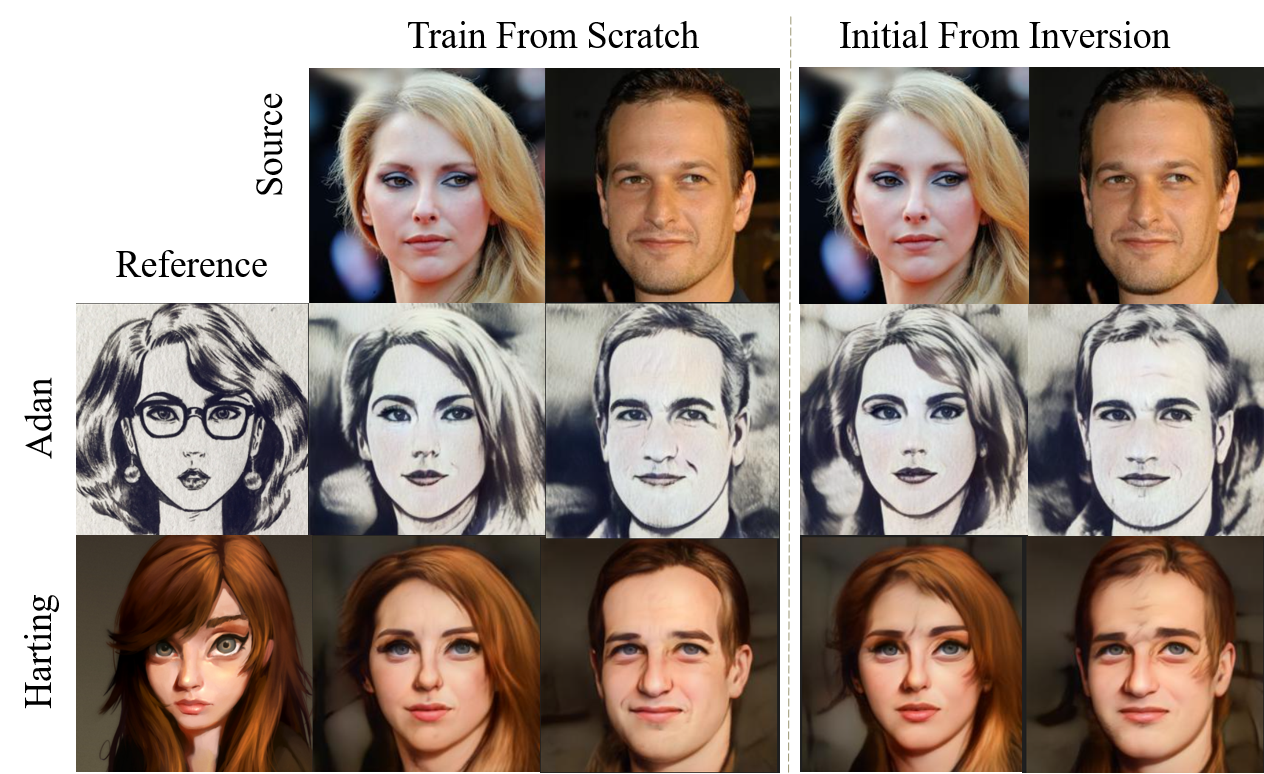}
\caption{Qualitative comparison of initialization for one-shot domain adaption.} \label{fig1}
\end{figure}

We use the StyleGAN2 pre-trained on FFHQ dataset in domain adaptation. For the text-driven domain adaptation model, we apply the pre-trained CLIP models 'ViT-B/16' and 'ViT-B/32' to calculate the direction loss following StyleGAN-NADA. We use the Adam optimizer to train the model for $300$ iterations with a batch size of 2 and a learning rate of $1 \times 10^{-3}$. For the one-shot domain adaptation model, we follow the default settings of GOGA. Additionally, we can utilize our two-stage model to extract the approximate token sets $P,Q$ of the reference image and provide an initialization for domain adaption. As shown in Fig.~\ref{fig1}, under the same iterations, our RefineStyle trained from inverted $P$ and $Q$ produces more stylized synthesis. All experiments are implemented on 1 NVIDIA RTX 3090 GPU.
\end{document}